\documentclass[10pt,journal]{IEEEtran}

\ifCLASSINFOpdf
\else
\fi
\usepackage{times}
\usepackage{epsfig}
\usepackage{graphicx}
\usepackage{amsmath}
\usepackage{amssymb}

\usepackage{footnote}
\usepackage{multirow}
\usepackage{booktabs}
\usepackage{threeparttable}
\usepackage{enumerate}
\usepackage{indentfirst}
\usepackage{array}
\usepackage{rotating}
\usepackage{ragged2e}
\usepackage{wasysym}
\usepackage{caption}
\usepackage{epstopdf}

\def\eg{{\em e.g.}}
\def\etal{{\em et al.}}

\newcommand{\figref}[1]{Fig. \ref{#1}}
\newcommand{\tabref}[1]{Tab. \ref{#1}}

\newcommand{\secref}[1]{Section \ref{#1}}

\newcommand{\myPara}[1]{\vspace{.05in}\noindent\textbf{#1}}

\newcommand{\bl}[1]{\textbf{#1}}
\newcommand{\mc}[1]{\mathcal{#1}}
\newcommand{\mb}[1]{\mathbb{#1}}
\newcommand{\tabincell}[2]{\begin{tabular}{@{}#1@{}}#2\end{tabular}}

\begin{document}
%
\title{Spatiotemporal Knowledge Distillation for Efficient Estimation of Aerial Video Saliency}
%
%
%

\author{Jia~Li,~\IEEEmembership{Senior Member,~IEEE},
        Kui~Fu,
        Shengwei~Zhao and~Shiming~Ge,~\IEEEmembership{Senior Member,~IEEE}
\IEEEcompsocitemizethanks{\IEEEcompsocthanksitem J. Li and K. Fu are with the State Key Laboratory of Virtual Reality Technology and Systems, School of Computer Science and Engineering, Beihang University, Beijing 100191, China. \protect
\IEEEcompsocthanksitem J. Li is also with the Beijing Advanced Innovation Center for Big Data and Brain Computing, Beihang University, Beijing, 100191, China, and the Peng Cheng Laboratory, Shenzhen, 518000, China. \protect
\IEEEcompsocthanksitem S. Zhao and S. Ge are with the Institute of Information Engineering, Chinese Academy of Sciences, Beijing, 100095, China. \protect
\IEEEcompsocthanksitem S. Zhao is also with the School of Cyber Security at University of Chinese Academy of Sciences, Beijing, 100095, China. \protect
\IEEEcompsocthanksitem S. Ge is the corresponding author. E-mail: geshiming@iie.ac.cn. \protect}
}

\maketitle

\begin{abstract}
  The performance of video saliency estimation techniques has achieved significant advances along with the rapid development of Convolutional Neural Networks (CNNs). However, devices like cameras and drones may have limited computational capability and storage space so that the direct deployment of complex deep saliency models becomes infeasible. To address this problem, this paper proposes a dynamic saliency estimation approach for aerial videos via spatiotemporal knowledge distillation. In this approach, five components are involved, including two teachers, two students and the desired spatiotemporal model. The knowledge of spatial and temporal saliency is first separately transferred from the two complex and redundant teachers to their simple and compact students, and the input scenes are also degraded from high-resolution to low-resolution to remove the probable data redundancy so as to greatly speed up the feature extraction process. After that, the desired spatiotemporal model is further trained by distilling and encoding the spatial and temporal saliency knowledge of two students into a unified network. In this manner, the inter-model redundancy can be further removed for the effective estimation of dynamic saliency on aerial videos. Experimental results show that the proposed approach outperforms ten state-of-the-art models in estimating visual saliency on aerial videos, while its speed reaches up to 28,738 FPS on the GPU platform.
\end{abstract}


%
\IEEEpeerreviewmaketitle

\section{Introduction}\label{sec:intro}

\IEEEPARstart{T}{he} rapid development of mobile devices further emphasizes the importance of effectively and efficiently estimating dynamic visual saliency on videos. For example, a drone, one of the most popular mobile devices in recent years, is capable of collecting high-resolution aerial videos in various scenarios due to its flexible operability. To analyze these high-resolution videos on the drone with limited memory and computational capability, a highly efficient and accurate saliency model is required so that the limited resources can be spent on the attractive visual content with a high priority. In this manner, subsequent complex vision tasks such as event understanding \cite{shu2015joint} and drone navigation \cite{zhang2010novel} can be widely facilitated in both speed and accuracy.

\begin{figure}[t]
\begin{center}
   \includegraphics[width=1.0\columnwidth]{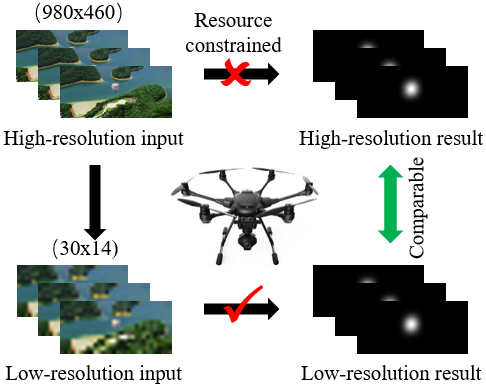}
\end{center}
   \caption{Low-resolution frames lack details but salient targets can still be easily localized by the human-being. Since the directly deployment of complex saliency models on resource-limited drones may have difficulty in processing high-resolution aerial videos, a feasible solution is to distil the knowledge from complex models into a simple and compact model and remove the data redundancy (\eg, resolution degradation) to ensure highly efficient aerial video processing.}
\label{fig:motivation}
\end{figure}

In the past decades, many models have been constructed in visual saliency estimation by defining comprehensive rules \cite{Goferman2012Context, Li2015Finding, zhang2018saliency, ding2018improving} or using deep learning frameworks \cite{Pan2016Shallow, wang2018deep, bak2017spatio}. In particular, the deep models have achieved impressive performance along with the development of large-scale benchmark datasets~\cite{deng2009imagenet, CAT2000, abu2016youtube} but at the cost of huge memories and massive computations. However, these ``ground-level'' deep models may have difficulties to be directly deployed on drones for processing high-resolution aerial videos. The main reasons are two-fold: 1)~the limited computational resource on drones is far from sufficient to meet the requirement of complex deep models; and 2)~the ground-level saliency models may have difficulties to handle aerial videos since the data distributions change remarkably. As a result, to deploy these complex saliency models on resource-limited drones, two issues should be addressed first: 1)~How to reduce the computational cost and memory footprint of deep saliency models without remarkable loss of accuracy? and 2)~How to fuse both spatial and temporal cues to extract powerful features that apply to aerial videos?

\begin{figure*}[t]
\begin{center}
   \includegraphics[width=1.0\textwidth]{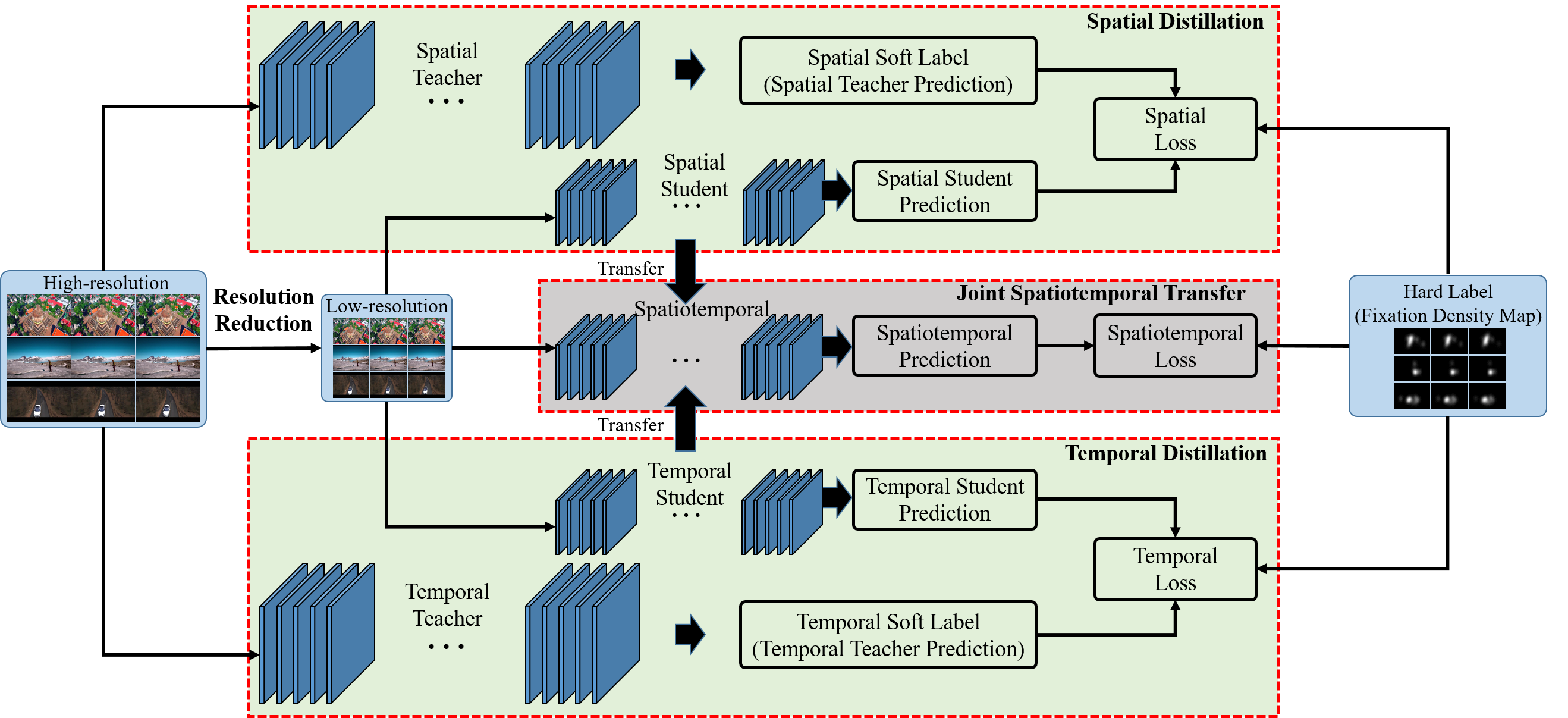}
\end{center}
   \caption{System framework. The framework consists of five components: two teachers, two students and the desired spatiotemporal model. The knowledge is transferred from teachers to the desired model via two steps: 1)~distilling the knowledge separately from the spatial teacher $\mb{T}_s$ and the temporal teacher $\mb{T}_t$ to their students $\mb{S}_s$ and $\mb{S}_t$, respectively. The distillation is conducted along with resolution degradation to remove data redundancy. 2)~transferring and fusing the knowledge of $\mb{S}_s$ and $\mb{S}_t$ into a unified spatiotemporal model $\mb{S}_{st}$ to improve the accuracy and speed of dynamic saliency estimation on aerial videos.}
\label{fig:framework}
\end{figure*}

To address these issues, we inspect existing deep models and find two major factors that restrict the computational cost: the model redundancy and the data redundancy. Typically, some researchers tend to train deeper networks from large-scale datasets for dynamic saliency estimation. However, the models may become highly redundant for such a low-level vision task, leading to the dramatic increase of computation and memory costs. This fact facilitates the studies of model compression \cite{denil2013predicting, kim2015compression, zhang2016accelerating, zhang2015efficient} that aim to remove the redundancy to generate compact models by parameter pruning and quantization. In this way, the resulting compact models usually have low computational cost and memory footprint but often encounter a sharp accuracy drop. To address this issue, some works \cite{bucilu¨£2006model, hinton2015distilling, romero2014fitnets} proposed to distil the knowledge from complex models and then transfer to simple ones without a significant performance drop.

Beyond the model redundancy, the data redundancy is less considered. Actually, visual details are not always necessary especially in low-level vision tasks such as visual saliency estimation. As shown in \figref{fig:motivation}, salient targets in the heavily blurred low-resolution frames can still be easily localized by the human-being without such details. This implies that there exist strong data redundancy that are not necessary for the saliency estimation task. As a consequence, removing such data redundancy may be another way to further reduce the computational cost.

Inspired by these two findings, this paper proposes a spatiotemporal knowledge distillation approach. As shown in \figref{fig:framework}, the framework of this approach consists of five components, including two teachers, two students and the desired spatiotemporal model. The knowledge of visual saliency is transferred from the spatial and temporal teachers to the final spatiotemporal model by using the spatial and temporal students as the bridges. In this process, the spatial and temporal knowledge is first separately extracted from complex and redundant teachers and then transferred into simple and compact students to remove intra-model redundancy. Meanwhile, the input scenes are also degraded from high-resolution to low-resolution to remove data redundancy. After that, the desired spatiotemporal model is trained by distilling and encoding the spatial and temporal knowledge of the two students into a unified network to further remove the inter-model redundancy. By step-wisely removing the intra-model, data and inter-model redundancies, the dynamic saliency of aerial videos can be effectively estimated with an extremely high speed. Experimental results show that the proposed approach outperforms ten state-of-the-art models. In particular, its speed can reach up to 28,738 FPS on the GPU platform.

Our main contributions are summarized as follows: 1) We propose a two-step knowledge distillation framework that can greatly reduce the computational cost with little accuracy drop in aerial video saliency estimation; 2) We design a lightweight spatiotemporal network that can extract and fuse both spatial and temporal saliency cues; and 3) we conduct extensive experiments and prove that our approach achieves an ultrafast speed and outperforms ten state-of-the-art models.

The rest of this paper is organized as follows: \secref{sec:related} reviews related works and \secref{sec:method} presents the spatiotemporal knowledge distillation. \secref{sec:exp} benchmarks the proposed model. Finally, \secref{sec:conclusion} concludes the paper.
\section{Related Works}
\label{sec:related}
In this paper, we aim to distil knowledge from well pretrained saliency models that serve as teachers and transfer their knowledge to the students to facilitate efficient saliency estimation. Therefore, we present a brief review of visual saliency models and knowledge distillation studies.

\subsection{Visual Saliency Models}
The recent advances in the field of saliency estimation from videos result in many visual saliency models. These models can be roughly grouped into three categories according to their features and frameworks, including heuristic, shallow-learning and deep-learning saliency models.

The heuristic saliency models \cite{itti1998model, itti2004automatic, seo2009static, jiang2013salient, li2014saliency} generally use hand-crafted features and design heuristic rules to perform visual saliency estimation in a bottom-up or top-down manner. The bottom-up models are stimulus-driven and compete fairly to pop-out conspicuous visual signals. In these models, hand-crafted features such as directions, colors and intensities as well as heuristic fusion rules are widely used. For example, Fang~\etal~\cite{fang2014video} proposed a spatiotemporal framework to separately detect the spatial and temporal saliency cues. These cues were then fused according to the spatial compactness and the temporal motion contrast. Later, they~\etal~\cite{fang2017visual} proposed uncertainty weighting to fuse the spatial and temporal saliency results. However, such unbiased heuristic fusion strategies may have difficulties in suppressing background distractors. To alleviate this issue, some task-driven models heuristically incorporate high-level factors in a top-down manner. For example, Borji \etal~\cite{borji2012probabilistic} modeled the task-driven visual attention with a unified Bayesian approach by integrating global scene context, previous attention locations and motion actions to predict the next attention locations. Chen \etal~\cite{chen2016video} predicted video saliency by combining the bottom-up saliency maps and the top-down ones through point-wise multiplication. Generally speaking, these heuristic saliency models often perform efficiently in estimating saliency but may suffer from poor accuracy and low robustness since the hand-crafted features and heuristic fusion strategies may be not optimal for all scenarios.

Inspired by the pros and cons of heuristic models, the shallow-learning saliency models \cite{li2010probabilistic, lee2011learning, vig2012intrinsic, Vig2014Large} aim to directly learn an optimal fusion strategy of hand-crafted features from data. For example, the saliency model proposed by Vig \etal \cite{vig2012intrinsic} used supervised learning to fine-tune the free parameters in dynamic scenarios. Fang \etal~\cite{Fang2017Learning} proposed an optimization framework with pairwise binary terms by learning a set of discriminative subspaces to pop out targets and suppress distractors. Moreover, some works \cite{lee2011learning, song2014low} proposed to apply learning algorithms to combine multi-level features into the saliency estimation processes. For example, Song \etal~\cite{song2014low} estimated saliency by fusing the low-level and high-level features as well as the center-bias priors. Due to optimized fusion strategy, learning-based saliency models achieved promising results. However, these models inherently share an upper bound in the performance since the hand-crafted features used may be also not optimal.

To address the feature issue, deep saliency models \cite{hu2017deep, kuen2016recurrent, li2016deepsaliency} proposed to use feature representations learned from data by using Convolutional Neural Networks (CNNs). Some of these models directly employ the state-of-the-art deep models pretrained in large-scale visual tasks as feature extractors. For example, K\"{u}mmerer \etal reused AlexNet~\cite{krizhevsky2012imagenet} and VGG-19 \cite{simonyan2014very} to generate high-dimensional features for fixation prediction in \cite{kummerer2014deep} and \cite{kummerer2016deepgaze}, respectively. In contrast, Pan \etal \cite{Pan2016Shallow} proposed to train a shallow CNN and a deep CNN in an end-to-end manner for fixation prediction. In addition, some deep saliency models focus on designing specific architectures or loss functions. For example, Imamoglu \etal~\cite{imamouglu2017saliency} utilized the objectiveness scores predicted by the features selected from CNNs to detect conspicuous regions. Due to the rich knowledge extracted by the complex deep saliency models, the deep models usually outperform  heuristic and shallow-learning models in accuracy. However, the computational cost often increase remarkably due to the rich redundancy in both models (\eg, unnecessary computations) and data (\eg, unnecessary high-resolution inputs), which prevent them from being directly deployed on mobile devices such as drones and cameras. Therefore, it is necessary to compress or distil these saliency models to greatly reduce the computational cost without remarkable performance drop.

\subsection{Knowledge Distillation}
Knowledge distillation~\cite{hinton2015distilling} is a specific model compression technique that distils the inherent knowledge from a complex teacher model to a simple student one so as to greatly reduce the model redundancy and maintain a comparable performance. To this end, the student models are trained under the supervision of the teacher models in many existing works. For example, Hinton \etal~\cite{hinton2015distilling} introduced the soft labels generated by a teacher model as an extra supervision, which was combined with the hard supervision defined by data labels. There also exist some other forms of supervision such as classification probabilities \cite{hinton2015distilling}, feature representations~\cite{ba2014deep, romero2014fitnets}, and inter-layer flows (the inner product of feature maps) \cite{yim2017gift}. Zhang \etal~\cite{zhang2017deep} proposed deep mutual learning, which conducted online distillation in one-phase training between two peer student models. Rusu \etal~\cite{rusu2015policy} proposed a multi-teacher single-student policy to distil knowledge from multiple teachers into a single student.

Generally specking, these knowledge distillation approaches provide a powerful way to reduce the model redundancy, which is efficient in dealing with high-resolution static saliency estimation. However, the data redundancy is less considered, especially in low-level vision tasks like visual saliency estimation. Actually, saliency estimation is a low-level vision task that does not need so many details represented by high-resolution frame sequences. By removing such data redundancy hidden in the high-resolution as well as the consecutive video frames, the speed of a dynamic saliency model can be greatly boosted. To this end, we propose a spatiotemporal knowledge distillation approach to simultaneously reduce both the model and data redundancies while the model accuracy can be well maintained.

\section{The Proposed Approach} \label{sec:method}
In this section, we present a Spatiotemporal Knowledge Distillation (SKD) approach for dynamic saliency estimation in aerial videos. The proposed approach operates in two major steps: the separate spatial/temporal knowledge distillation, and the joint spatiotemporal knowledge transfer. Here we start with a brief overview of the approach and then elaborate on these two steps as well as their implementation details.

\subsection{The framework}
As shown in \figref{fig:framework}, the proposed SKD approach consists of five major components, including a spatial teacher, a temporal teacher, a spatial student, a temporal student and the desired spatiotemporal model. Note that the spatial/temporal teachers can be any complex spatial/temporal saliency models pretrained on massive high-resolution data. Typically, such teacher models give impressive performances in predicting spatial saliency (\eg, DVA~\cite{wang2018deep}, SalNet~\cite{Pan2016Shallow} and SSNet~\cite{bak2017spatio}) or temporal saliency (\eg, TSNet \cite{bak2017spatio}) at the expense of high computational cost. As a result, the objective of the proposed distillation framework is to distil their knowledge into much simpler student models and finally the desired spatiotemporal saliency model by removing the model and data redundancy in two consecutive steps.

In the first step, we separately distil the knowledge from spatial and temporal teachers to the two students to reduce intra-model redundancy, respectively. Meanwhile, the high-resolution inputs, which often contain unnecessary details for low-level vision tasks such as saliency estimation, are degraded into low-resolution ones to remove data redundancy.

In the second step, the knowledge in the spatial and temporal students is jointly transferred and encoded into the desired spatiotemporal model. In this manner, the inter-model redundancy of two students in extracting common visual features can be also removed. As a result, the intra-model, inter-model and data redundancies are step-wisely removed, leading to a model with high accuracy and extremely low computational cost.

\subsection{Separate Spatial and Temporal Knowledge Distillation}
The separate spatial and temporal distillation operations in the first step force two simple students to mimic the behavior of two complex teachers in a spatial high-resolution frame and a temporal consecutive frame pair, respectively. Let $\mc{D}=\{I_n, Y_n\}_{n=1}^{N}$ be the training dataset and $I_n$ be an high-resolution frame with the ground-truth saliency map $Y_n$. Then, we can easily compute their resolution-degraded version as $\hat{\mc{D}}=\{\hat{I_n}, \hat{Y_n}\}_{n=1}^{N}$ by using a resolution-reduction operation $\mc{R}$ so that $\hat{I_n}=\mc{R}(I_n)$ and $\hat{Y_n}=\mc{R}(Y_n)$. For the sake of simplification, we denote the spatial and temporal teachers as $\mb{T}_s$ and $\mb{T}_t$, respectively. Similarly, the spatial and temporal students are denoted as $\mb{S}_s$ and $\mb{S}_t$, respectively.
Note that $\mb{T}_s$ and $\mb{S}_s$ take a single frame as the input, while $\mb{T}_t$ and $\mb{S}_t$ use a pair of consecutive frames.

\begin{figure}[t]
\begin{center}
   \includegraphics[width=1.0\columnwidth]{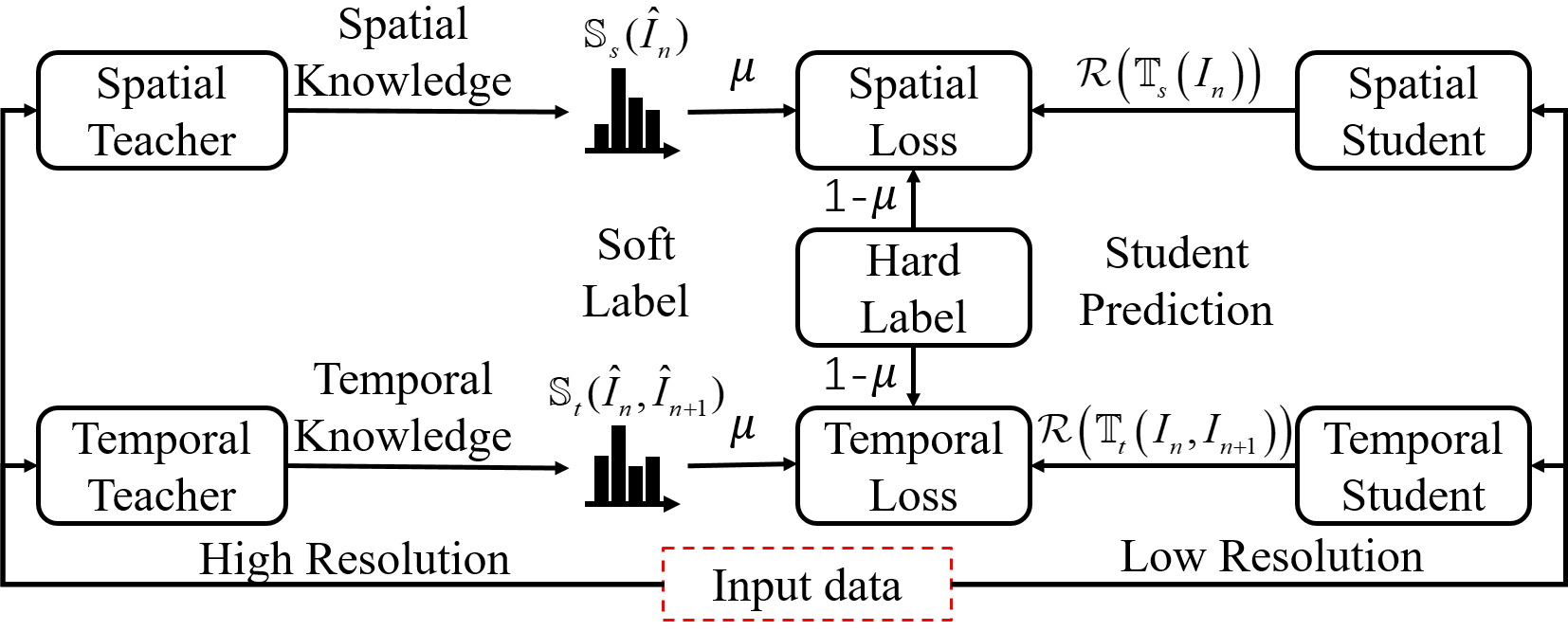}
\end{center}
   \caption{Distillation flow. The spatial/temporal student networks are trained under the supervision of hard labels as well as soft labels generated by spatial/temporal teacher networks. By this way, the private knowledge inherited in the spatial and temporal teacher networks can be transfer into the spatial and temporal student networks.}
\label{fig:distillation}
\end{figure}

Inspired by \cite{hinton2015distilling}, we first generate the high-resolution soft labels of teachers and use their resolution-degraded versions to supervise the training process of the two students. In this sense, the spatial and temporal students are trained to optimize the following spatial and temporal losses, respectively.
\begin{equation}
\label{eq:l_s}
\begin{split}
\mc{L}_s=&\mu\cdot\mc{L}_{soft}\left(\mb{S}_s(\hat{I}_n),\mc{R}(\mb{T}_s(I_n))\right)\\
+&(1-\mu)\cdot\mc{L}_{hard}\left(\mb{S}_s(\hat{I}_n),\hat{Y}_n\right),
\end{split}
\end{equation}
\begin{equation}
\label{eq:l_t}
\begin{split}
\mc{L}_t=&\mu\cdot\mc{L}_{soft}\left(\mb{S}_t(\hat{I}_n,\hat{I}_{n+1}),\mc{R}\left(\mb{T}_t\left(I_n,I_{n+1}\right)\right)\right) \\
&+(1-\mu)\cdot\mc{L}_{hard}\left(\mb{S}_t(\hat{I}_n,\hat{I}_{n+1}),\hat{Y}_n\right),
\end{split}
\end{equation}
where the scale parameter $\mu$ is used to balance the soft loss $\mc{L}_{soft}$ and hard loss $\mc{L}_{hard}$ (we empirically set $\mu=0.5$). The $\mc{L}_{soft}$ is used to measure the difference between the resolution-degraded predictions of teachers and their students, while $\mc{L}_{hard}$ is computed between the resolution-degraded ground-truth maps and the student predictions. Both of the two losses use normalized $\mc{L}_2$ loss:
\begin{eqnarray}
\mc{L}_{soft}(\mb{S},\mb{T})&=&\frac{1}{w\cdot{}h}\cdot{}\|\mb{S}-\mc{R}(\mb{T})\|^2_2, \\
\mc{L}_{hard}(\mb{S},Y)&=&\frac{1}{w\cdot{}h}\cdot{}\|\mb{S}-\mc{R}(Y)\|^2_2,
\end{eqnarray}
where $\mb{S}$ and $\mb{T}$ denote the predictions of spatial/temporal student and teacher, respectively. The $w$ and $h$ are the width and height of low-resolution video frames. The distillation flow in the first step is shown in \figref{fig:distillation}, where the high-resolution teacher knowledge is distilled into low-resolution students. We can see that this distillation flow seeks a balance of generalization ability and prediction accuracy. The soft labels given by the complex teacher models pre-trained on large-scale public or private datasets reflect a probabilistic understanding of the input scenes. With the supervision of the soft labels, the generalization ability of the student models can be enhanced. In this way, the computational cost can be greatly reduced without remarkable loss of accuracy.

\begin{figure}[t]
\begin{center}
   \includegraphics[width=1.0\columnwidth]{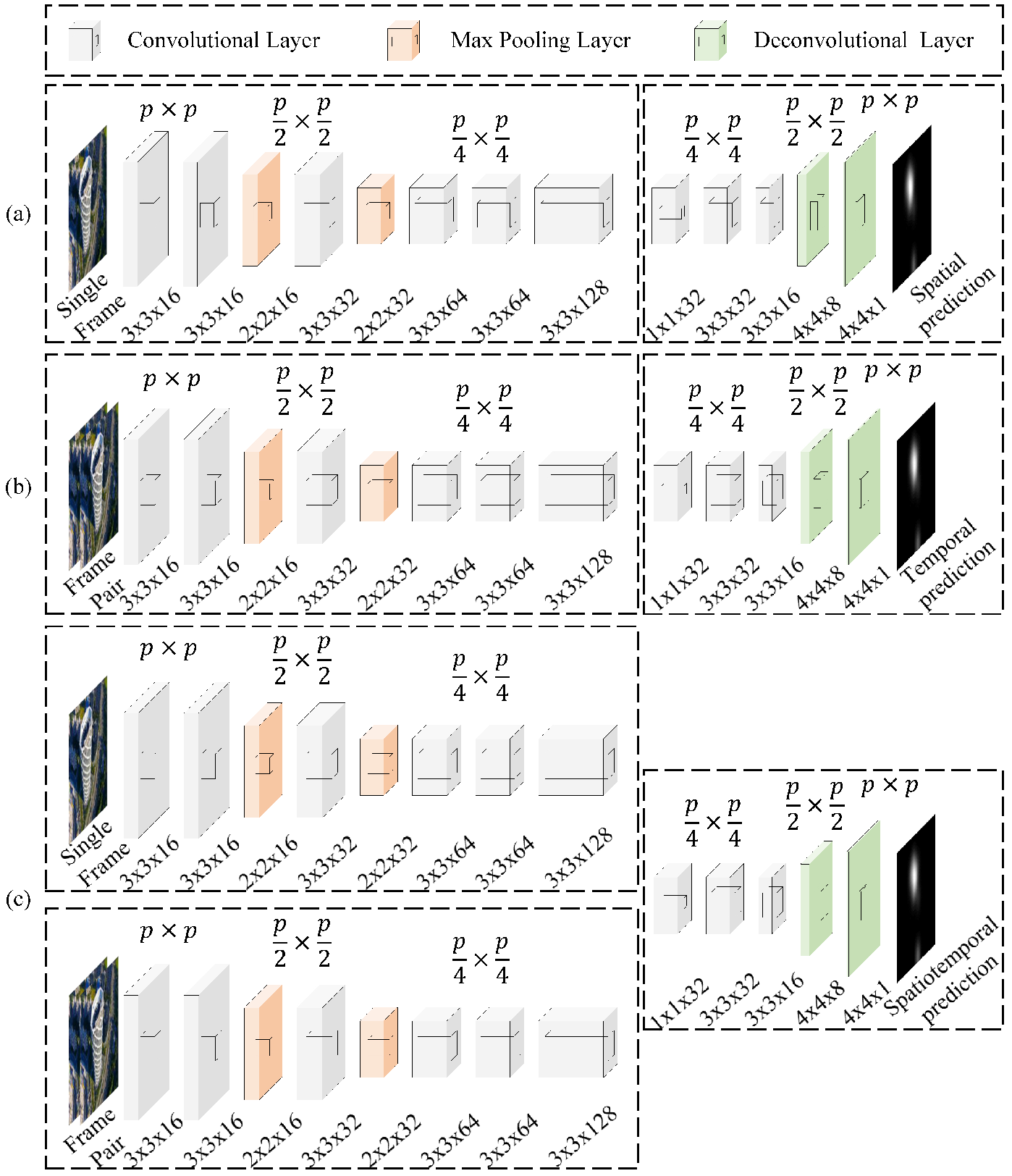}
\end{center}
   \caption{Networks. (a) Spatial student network. (b) Temporal student network. (c) Spatiotemporal network. The spatial network takes a single frame as the input, while the temporal network takes a pair of successive frames as the input.}
\label{fig:network}
\end{figure}

The detail structures of students are shown in \figref{fig:network} (a) and (b). The $\mb{S}_{s}$ is a fully convolutional network (FCN) which takes a single low-resolution frame $I_n$ as input. Inspired by SalNet \cite{Pan2016Shallow}, $\mb{S}_{s}$ contains $13$ layers. Aiming at dealing the small targets in aerial videos, we use the majority $3\times 3$ convolutional kernels to enhance the local information extraction ability. In order to gradually expand the receptive fields, we adopt two pooling layers, in the 3rd and 5th layers of each path, respectively. Convolutional layers with $1\times 1$ kernel size are adopted in the 9th layer to reduce the dimension of the feature maps while maintaining the diversity and effectiveness of the feature maps. A Rectified Linear Unit (ReLU) layer is adopted after every convolutional layer to improve feature representation capability. In this manner, we can obtain a low-level and mid-level feature extractor with good performance. After that, we use two convolutional layers with kernel size $3\times 3$ to extract high-level saliency cues. In addition, we design a decoder network which contains two deconvolutional layers to upsample feature maps and constructs an output that maintains the original resolution of the input.

The structure of $\mb{S}_t$ is similar to $\mb{S}_s$ but has difference in the 1st layer. The $\mb{S}_t$ takes a pair of successive low-resolution frames $(\hat{I}_n,\hat{I}_{n+1})$ as the input to directly capture the temporal cues in aerial videos. In practice, we concatenate the current frame $\hat{I}_n$ and the next frame $\hat{I}_{n+1}$ to an input tensor with the size of $h\times w\times 6$. Note that the teacher models can be any classic deep models trained in existing public or private datasets, and we fine-tune them on the high-resolution aerial videos so that they can adapt to the specific visual attributes of aerial videos such as large-scale scenarios, small targets and vertical viewpoints.

\subsection{Joint Spatiotemporal Knowledge Transfer}
After the separate spatial and temporal knowledge distillation, the teacher knowledge has been distilled into the corresponding students. Considering that the spatial student takes one frame as the input while the temporal student takes a pair of frames, there surely exist some redundancy in these two student models, especially in the feature extraction. To further remove such inter-model redundancy, we conduct a joint spatiotemporal knowledge transfer step to extract compact and powerful spatiotemporal saliency features with the desired spatiotemporal model $\mb{S}_{st}$.

The network architecture of $\mb{S}_{st}$ is shown in \figref{fig:network}(c), which has two input information streams. The spatial and temporal input streams share the same structure as the first eight layers of the spatial and temporal students to extract the spatial features $\mc{F}_s$ and the temporal features $\mc{F}_t$, respectively. After that, these two streams are combined into a fusion sub-network which takes a similar structure as the last four layers to the students. The input $\mc{F}$ of the fusion sub-network is the concatenation of $\mc{F}_s$ and $\mc{F}_t$.

\begin{figure}[t]
\begin{center}
   \includegraphics[width=1.0\columnwidth]{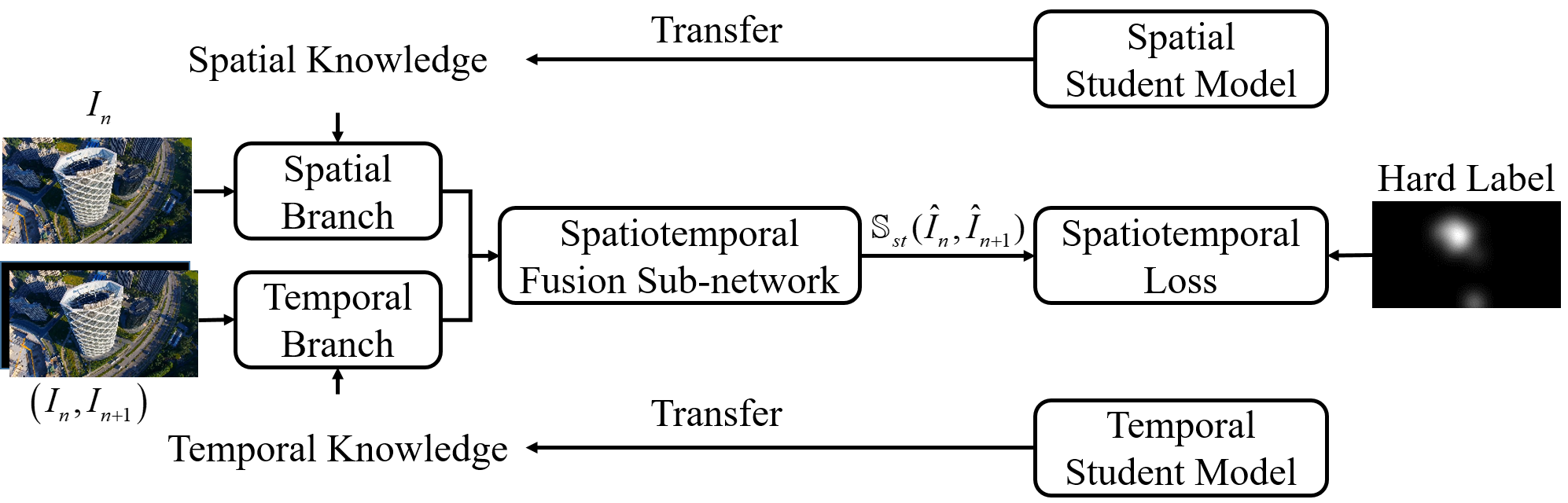}
\end{center}
   \caption{The knowledge transfer process in training the desired spatiotemporal saliency model. The spatial and temporal knowledge learned by the spatial and temporal student networks is transferred into the spatiotemporal network for extracting spatial feature $\mc{F}_s$ and temporal feature $\mc{F}_t$. Then the spatiotemporal network fuses them for extracting powerful spatiotemporal featuresp for better performance.}
\label{fig:optimization}
\end{figure}

During training $\mb{S}_{st}$, we initialize its first eight layers with the spatial and temporal student models and the fusion sub-network with a truncated random normal distribution. Then the training process is performed by optimizing the following hard spatiotemporal loss $\mc{L}_{st}$:
\begin{equation}
\mc{L}_{st}=\mc{L}_{hard}(\mb{S}_{st}(\hat{I}_n,\hat{I}_{n+1}),\hat{Y_n}).
\end{equation}
The knowledge transfer process in training the desired spatiotemporal model is shown in \figref{fig:optimization}. We can see that the knowledge is first transferred from the two students into the two streams of the desired spatiotemporal model, and the features extracted by the two streams are further fine-tuned on low-resolution aerial videos in a fully supervised manner to remove the redundancy in the spatial and temporal students.

In the implementation, the model adopts the Tensorflow platform \cite{abadi2016tensorflow} on NVIDIA GPU 1080Ti and a single core Intel CPU 3.4GHz. The learning rate and batch size are set as $1\times 10^{-3}$ and $128$, respectively. The Optimizer adopts Adam algorithm \cite{kingma2014adam}. After training, the learned spatiotemporal model is deployed for processing aerial videos collected by drones. It takes a successive low-resolution frame pair $(\hat{I}_n,\hat{I}_{n+1})$ as the input. Then, the spatial stream receives $\hat{I}_n$ to generate spatial features, while the temporal branch takes $(\hat{I}_n,\hat{I}_{n+1})$ to output the temporal features.

\section{Experiments}
\label{sec:exp}

To verify the effectiveness and efficiency of the proposed SKD approach, we conduct the experiments on a large-scale aerial video benchmark dataset. We first introduce the experimental settings and then benchmark with ten state-of-the-art models. Finally, we conduct several diagnostics experiments to give an insight analyze of SKD approach.


\subsection{Experimental Setting}

All experiments are conducted on AVS1K \cite{fu2018drones}, a largest aerial video dataset for saliency estimation. AVS1K contains $1,000$ aerial videos and $177,644$ frames. Its maximal video resolution and average video length are $1280\times 720$ and $5.92$s, respectively. According to the salient targets, the videos in AVS1K can be divided into four categories: Building (AVS1K-B), Human (AVS1K-H), Vehicle (AVS1K-V) and Others (AVS1K-O):
\begin{itemize}
  \item \bl{AVS1K-B} contains $240$ aerial videos with $41,471$ frames, and the average video length is is $5.76$s.
  \item \bl{AVS1K-H} contains $210$ aerial videos with $31,699$ frames, and the average video length is is $5.03$s.
  \item \bl{AVS1K-V} contains $200$ aerial videos with $27,092$ frames, and the average video length is is $4.52$s.
  \item \bl{AVS1K-O} contains $240$ aerial videos with $77,402$ frames, and the average video length is $7.37$s.
\end{itemize}

To evaluate the proposed approach, we quantitatively compare its performance against that of ten state-of-the-art models from three category groups:

\myPara{1)~The Heuristic Group (denoted as H Group)} contains three heuristic models, including HFT \cite{li2013visual}, SP \cite{li2014visual} and PNSP \cite{fang2014video}.

\myPara{2)~The Shallow-Learning Group (denoted as S Group)} contains two shallow-learning based models, including SSD \cite{Li2015Finding} and LDS \cite{Fang2017Learning}.

\myPara{3)~The Deep Learning Group (denoted as D Group)} contains five deep learning-based models, including eDN \cite{Vig2014Large}, iSEEL \cite{tavakoli2017exploiting}, DVA \cite{wang2018deep}, SalNet \cite{Pan2016Shallow} and STS \cite{bak2017spatio}.


Based on the investigation in \cite{riche2013saliency,li2015data,bylinskii2016different}, we report quantitative evaluation results on five widely used evaluation metrics, including the traditional Area Under the ROC Curve (AUC), the shuffled AUC (sAUC), the Normalized Scanpath Saliency (NSS), the Similarity Metric (SIM)~\cite{hou2013visual} and Correlation Coefficient (CC)~\cite{borji2012boosting}. AUC intuitively reflects the classification ability of ROC curve, which is generated by enumerating all probable thresholds of true positive rate versus false positive rate. Different from AUC, sAUC takes the fixations shuffled from other frames as negatives in generating the curve. NSS measures the average response at the eye fixation locations and normalizes the estimated saliency maps to zero mean and unit standard deviation. In this paper, the implementation in \cite{marat2009modelling} is adopted, which efficiently computes NSS via element-wise multiplication of the estimated and ground-truth saliency maps. SIM is computed to measure the similarity between the estimated and ground saliency maps, while CC is computed as the linear correlation between them. Noting that the values of all metric are positively correlated with the model performance. However, individual metric can not perfectly indicate whether the model is efficient or not. For example, AUC prefers to assign high score to a saliency map if it correctly predicts the order of saliency and less-salient locations, even if it is fuzzy. While sAUC and NSS trend to clean saliency maps that only pop-out the most salient locations and suppress all the distractors. Particularly, we take NSS as the primary metric according to surveys on saliency evaluation metrics~\cite{wang2018deep, liu2016deep}.

\subsection{Performance Evaluation on Aerial Videos}

\begin{figure}[t]
\begin{center}
   \includegraphics[width=1.0\columnwidth]{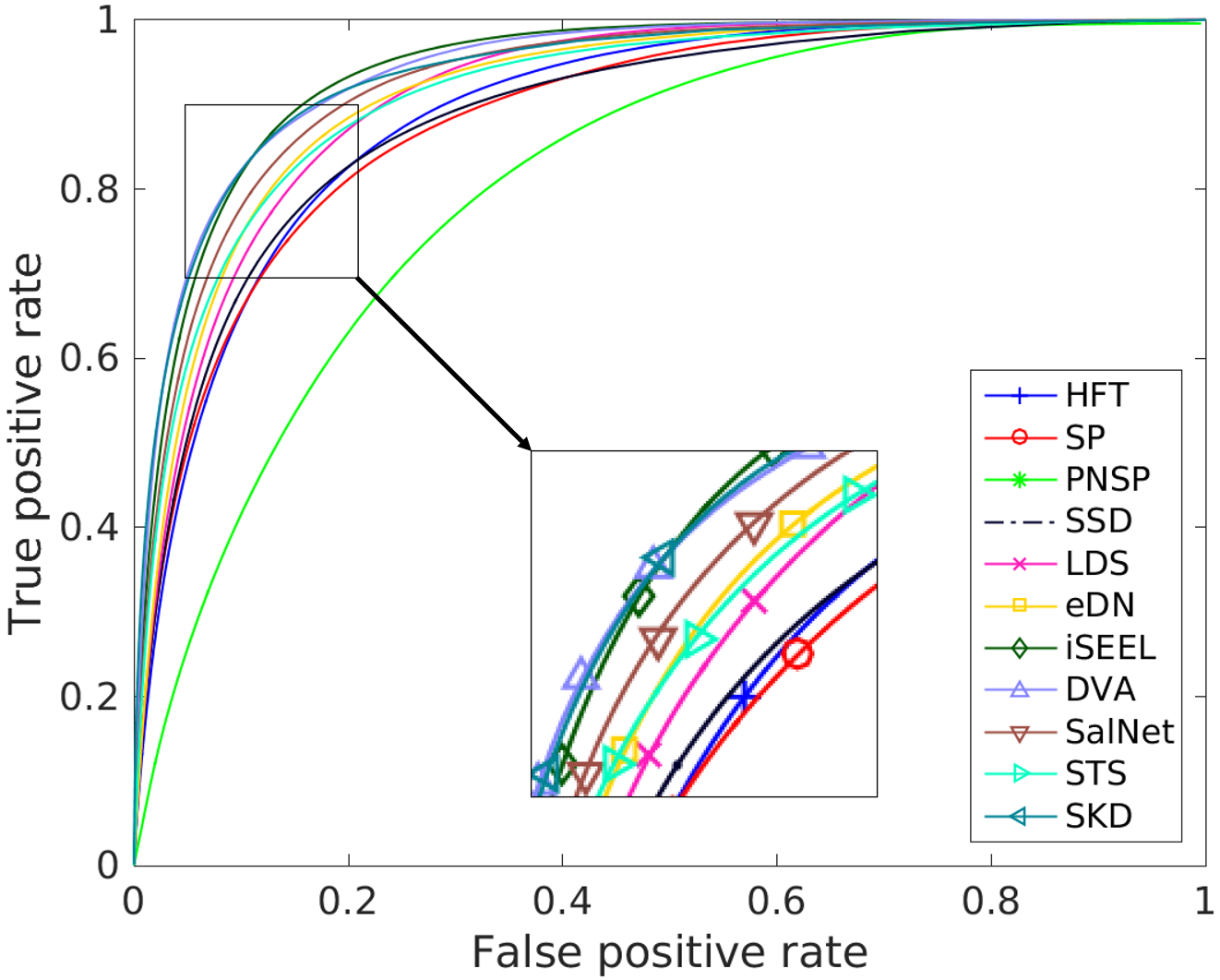}
\end{center}
   \caption{ROC curves of 11 models on AVS1K.}
\label{fig:ROC_AVS1K}
\end{figure}

\begin{figure*}[t]
\begin{center}
   \includegraphics[width=1.0\textwidth]{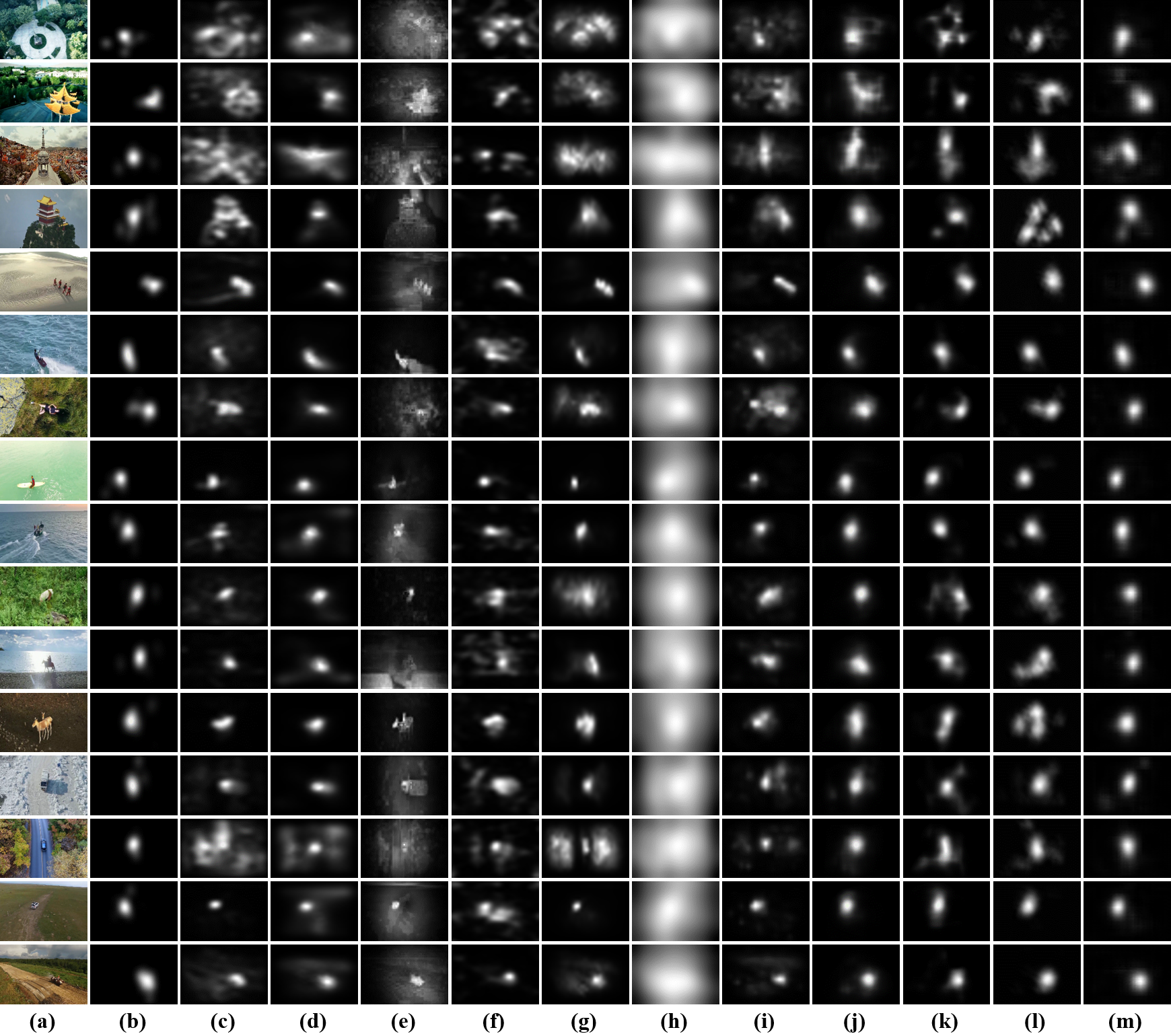}
\end{center}
   \caption{Representative frames of the models on AVS1K. (a) Video frame, (b) Ground truth, (c) HFT, (d) SP, (e) PNSP, (f) SSD, (g) LDS, (h) eDN, (i) iSEEL, (j) DVA, (k)  SalNet, (l) STS, (m) SKD.}
\label{fig:Result_AVS1K}
\end{figure*}

For simplicity, we use TSNet \cite{bak2016two} as the fixed temporal teacher and denote our models as {SKD}-$\mb{T}_s$-{R} where $\mb{T}_s$ is the spatial teacher model and {R} indicates the input resolution.
The performance of ten state-of-the-art and our two models on the AVS1K is presented in \tabref{tab:performance_AVS1K}. Here, {SKD-DVA-32} and {SKD-DVA-64} use DVA as the spatial teacher and take the input resolution of $32\times 32$ and $64\times 64$, respectively. Moreover, the ROC Curves are given in \figref{fig:ROC_AVS1K}. Some representative results of these models are shown in \figref{fig:Result_AVS1K}.
\begin{table*}[t]
\footnotesize
\centering
\caption{Performance comparison of ten state-of-the-art and our two models on AVS1K. The best and runner-up models of each column are marked with bold and underline, respectively. The models fine-tuned on AVS1K are marked with *.}
\label{tab:performance_AVS1K}
\begin{tabular}{l|l|ccccc|c|c|c|c}
\toprule
\multicolumn{2}{c|}{\multirow{3}{*}{Models}} &~\multirow{3}{*}{AUC}&~\multirow{3}{*}{sAUC}&~\multirow{3}{*}{NSS}&~\multirow{3}{*}{SIM}&~\multirow{3}{*}{CC} &~\multirow{3}{*}{\tabincell{c}{Parameters \\ (M)}} &~\multirow{3}{*}{\tabincell{c}{Memory \\ Footprint \\ (MB)}} &\multicolumn{2}{c}{Speed (FPS)} \\ \cline{10-11}
\multicolumn{2}{c|}{} & & & & & & & &~\tabincell{c}{GPU \\ (NVIDIA 1080Ti)}&~\tabincell{c}{CPU \\ (Intel 3.4GHz)} \tabularnewline
\midrule
\multirow{3}{*}{\bl{H}}&~HFT~\cite{li2013visual}            &~0.789 &~0.715 &~1.671 &~0.408 &~0.539 &~--- &~--- &~--- &~7.6  \tabularnewline
 &~SP~\cite{li2014visual}                                   &~0.781 &~0.706 &~1.602 &~0.422 &~0.520 &~--- &~--- &~--- &~3.6     \tabularnewline
 &~PNSP~\cite{fang2014video}                                &~0.787 &~0.634 &~1.140 &~0.321 &~0.370 &~--- &~--- &~--- &~---   \tabularnewline
\midrule
\multirow{2}{*}{\bl{S}}&~SSD~\cite{Li2015Finding}          &~0.737 &~0.692 &~1.564 &~0.404 &~0.503 &~--- &~--- &~--- &~32.2   \tabularnewline
 &~LDS~\cite{Fang2017Learning}                              &~0.808 &~0.720 &~1.743 &~0.452 &~0.565 &~--- &~--- &~--- &~4.6    \tabularnewline
\midrule
\multirow{8}{*}{\bl{D}}&~eDN~\cite{Vig2014Large}           &~0.855 &~0.732 &~1.262 &~0.289 &~0.417 &~--- &~--- &~--- &~0.2    \tabularnewline
 &~iSEEL~\cite{tavakoli2017exploiting}                      &~0.801 &~0.767 &~1.974 &~0.458 &~0.636 &~--- &~--- &~--- &~---   \tabularnewline
 &~DVA$^*$~\cite{wang2018deep}                              &~\underline{0.864} &~0.761 &~\underline{2.044} &~\bl{0.544} &~\underline{0.658} &~25.07 &~59.01 &~49 &~2.5    \tabularnewline
 &~SalNet$^*$~\cite{Pan2016Shallow}                         &~0.797 &~\underline{0.769} &~1.835 &~0.410 &~0.593 &~25.81 &~43.22 &~28 &~1.5     \tabularnewline
 &~STS$^*$~\cite{bak2017spatio}                             &~0.804 &~0.732 &~1.821 &~0.472 &~0.578 &~41.25 &~86.94 &~17 &~0.9     \tabularnewline
 &~\bl{SKD-DVA-32}                                          &~0.859 &~0.760 &~2.040&~0.527 &~0.657  &~\bl{0.30} &~\bl{0.14} &~\bl{28,738} &~\bl{1,490.5}   \tabularnewline
 &~\bl{SKD-DVA-64}                                          &~\bl{0.867} &~\bl{0.770} &~\bl{2.100} &~\underline{0.534} &~\bl{0.674}  &~\underline{0.30} &~\underline{0.58} &~\underline{8,522} &~\underline{411.7}   \tabularnewline
\bottomrule
\end{tabular}
\end{table*}

From \tabref{tab:performance_AVS1K}, we observe that the SKD-DVA-32 is comparable to the ten state-of-the-art models while the SKD-DVA-64 outperforms all models. Particularly, the SKD-DVA-64 ranks the first place in terms of AUC, sAUC, NSS and CC, but slightly worse than the DVA in term of SIM. Such performance improvement can be attributed to the spatiotemporal distillation framework. It distils the spatial and temporal knowledge inherited in the teachers into students in the separate spatial/temporal distillation step. Then the framework transfers such spatial and temporal knowledge into a desired spatiotemporal model and fine-tunes it for better estimation accuracy. Experimental results reveal that the SKD-DVA-64 has better representation capability when compared with traditional single stream networks (\eg, SalNet), classic two-stream networks for dynamical scenarios (\eg, STS) as well as multi-stream networks (\eg, DVA). In term of NSS, SKD-DVA-64 achieves $2.7$\%, $14.4$\% and $15.3$\% performance gain to DVA, SalNet and STS, respectively. It is worth to note that the continued decrease in resolution results in a performance attenuation to some extent. Intuitively, the SKD-DVA-32 has a $2.9$\% accuracy drop to SKD-DVA-64 in term of NSS.

We also find that the proposed approach can achieve an ultrafast speed in aerial video saliency estimation, which can be explained by the extremely low computational cost. Our spatiotemporal network has only $0.30$M parameters, namely with a $98.8$\% reduction to DVA. Benefiting from the combined effect of reduced parameters and input resolution, the computational cost and the memory footprint of the proposed approach are compressed into a extremely low extent. The SKD-DVA-32 and SKD-DVA-64 can achieve $421.5\times$ and $101.7\times$ memory reduction to DVA, respectively. In summary, the SKD-DVA-32 can achieve an ultrafast speed ($28,738$~FPS) with comparable performance to ten state-of-the-art models, while the SKD-DVA-64 can achieve a very fast speed ($8,522$~FPS) and outperforms ten state-of-the-art models in terms of most metrics.

Additionally, we can observe the difference lies among different categories. The heuristic models in the H group have the poorest performance. The reason may be that these heuristic models usually rely on low-level hand-crafted features and predefined rules for feature fusion. Thus these models may encounter huge challenges when infer saliency cues in unknown scenarios. BY adopting learnable fusion strategies, the models in S group can achieve slightly better performance but still far from satisfactory. The key issue is that the hand-crafted features adopted in H group and S group are designed for ground-level scenarios, which may not be applicable to aerial videos. This also indicates that there may exits some unconventional visual patterns in such aerial scenarios, which should be learned from the data. \tabref{tab:performance_AVS1K} reveals that the models in D group generally exceed that of the H group and S group, which can be attributed to the powerful capabilities of CNNs in extracting hierarchical feature representations.

\subsection{Diagnostics Experiments}
After the promising performance is achieved, we further conduct five diagnostics experiments on AVS1K to delve into our SKD framework. In the experiments, SKD-DVA-64 is taken as the baseline model.

\begin{figure}[t]
\begin{center}
   \includegraphics[width=1.0\columnwidth]{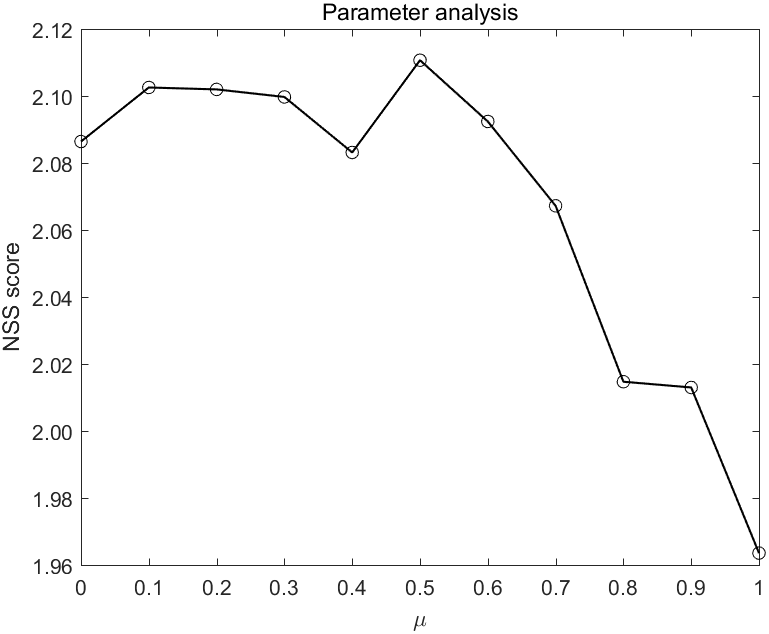}
\end{center}
   \caption{Parameter analysis on AVS1K with different $\mu$ in the interval [0.0, 1.0].}
\label{fig:NSS_curve}
\end{figure}

\myPara{Parameter Analysis.} In the first experiment, we analyze the parameter $\mu$ in \eqref{eq:l_s} and \eqref{eq:l_t} that is served as a scale parameter in computing $\mc{L}_s$ and $\mc{L}_t$. The curve of NSS scores on AVS1K with different $\mu$ is shown in \figref{fig:NSS_curve}, which is computed as the mean performance value in three tests. We find that the average NSS is relatively high (greater than 2.08) when $\mu$ falls between [0.0, 0.6]. Particularly, the desired spatiotemporal model achieves the best performance when the $\mu$ is set to 0.5. However, when the $\mu$ continues to grow, the performance drops sharply. This can be interpreted as it is difficult for soft labels to accurately represent the true distribution of data, and the supervision of hard labels is indispensable. In other words, the soft labels provide an opportunity to improve the generalization ability, while the hard labels emphasize only the accuracy. When both the generalization ability and accuracy are taken into consideration, the overall performance on the testing set can become better.

\myPara{Generalization Analysis.} The second experiment presents the performance of SKD-DVA-64 on four subsets to verify its generalization, as shown in \tabref{tab:performance_sub}. We can find that SKD-DVA-64 achieves better performance on AVS1K-H and AVS1K-V than that on AVS1K-B and AVS1K-O. The reason arises from that the targets like human and vehicles have appropriate sizes and conspicuous motion patterns on most aerial videos so that the spatiotemporal feature extraction can be easier. By contrast, AVS1K-B and AVS1K-O have relative static or larger targets, making the saliency model difficult to separate the targets from the distractors. This result implies that our model can generalize the ability in estimating the salient targets.
\begin{table}[t]
\centering{
\caption{Performance of SKD-DVA-64 on four subsets of AVS1K. The best and runner-up models of each column are marked with bold and underline, respectively.}
\label{tab:performance_sub}
\begin{tabular}{c|ccccc}
\toprule
Subset&~AUC&~sAUC&~NSS&~SIM&~CC \tabularnewline \midrule
{AVS1K-B}&~0.858&~0.762&~1.901&~0.535&~0.663 \tabularnewline
{AVS1K-H}&~\bl{0.903}&~\underline{0.795}&~\bl{2.465}&~\underline{0.551}&~\underline{0.719} \tabularnewline
{AVS1K-V}&~\underline{0.883}&~\bl{0.804}&~\underline{2.396}&~\bl{0.551}&~\bl{0.728} \tabularnewline
{AVS1K-O}&~0.849&~0.804&~1.934&~0.519&~0.637 \tabularnewline
\bottomrule
\end{tabular}
}
\end{table}

\begin{figure}[t]
\begin{center}
   \includegraphics[width=1.0\linewidth]{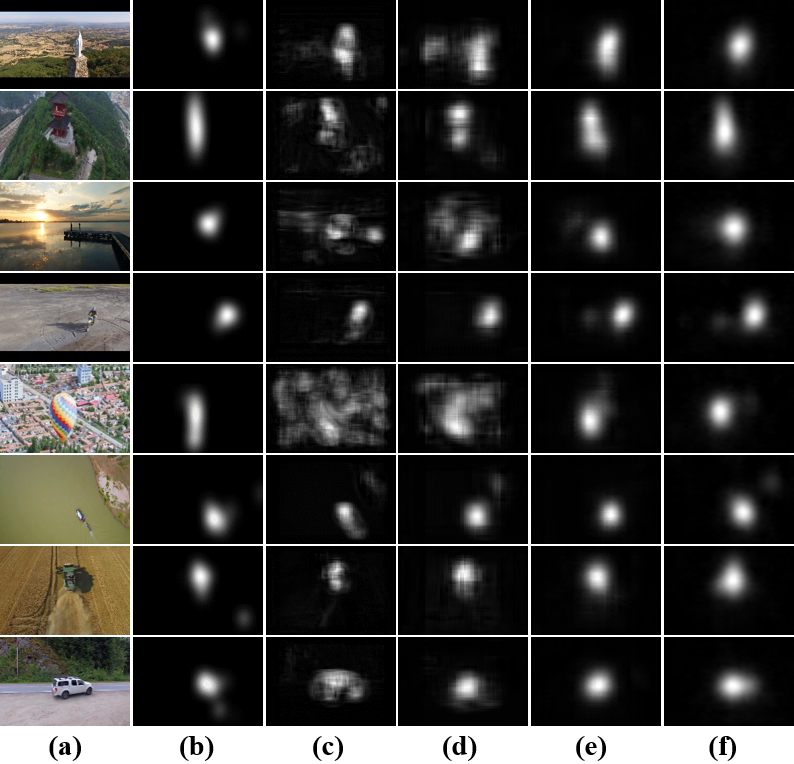}
\end{center}
   \caption{Representative frames of the proposed model in various resolutions on AVS1K. (a) Video frame, (b) Ground truth, (c) SKD-DVA-256, (d) SKD-DVA-128, (e) SKD-DVA-64, (f) SKD-DVA-32.}
\label{fig:Result_dis_AVS1K}
\end{figure}

\myPara{Teacher and Resolution.} The third experiment aims to assess the effect of the distilled teacher and input resolution. Without loss of generality, we fix the temporal teacher model as TSNet, and consider three candidate spatial teachers including DVA, SalNet and SSNet \cite{bak2016two}, that is $\mb{T}_s$$\in\{$DVA, SalNet, SSNet$\}$. Meanwhile, we also check four input resolutions, having R=$\{256, 128, 64, 32\}$. The performance of our different models is presented in \tabref{tab:distillation_AVS1K}, which shows some observations. First, under the same teacher, the input resolution has a remarkable effect on the model performance in terms of all metrics. The best and runner-up performance are achieved when the input resolutions are $64\times64$ and $32\times32$, respectively. It reveals that the data redundancy could be efficiently removed by using our framework, leading to better performance. Second, the performance is consistent under the same input resolution. For example, SKD-$\mb{T}_s$-64 models ranks top-1 no matter what spatial teacher model it adopts. It indicates that our framework provides a general way to distil teacher knowledge for improving saliency estimation. Third, the results generated by the models in a low resolution of $32\times32$ still have competitive performance. \figref{fig:Result_dis_AVS1K} shows some representative results, where the results tend to be clearer when the input resolution is reducing, which can be interpreted as the spatial network's receptive filed is small and insensitive to large-scale targets.

\begin{table}[t]
\footnotesize
\centering
\caption{The performance of the models under different settings on AVS1K dataset. ${T_S}$: spatial teacher model, Res: input resolution. The best and runner-up models of each column in each spatial teacher signal are marked with bold and underline, respectively.}
\label{tab:distillation_AVS1K}
\begin{tabular}{c|c|ccccc}
\toprule
${T_S}$ &Res     &~AUC&~sAUC&~NSS&~SIM&~CC \tabularnewline
\midrule
\multirow{4}{*}{\begin{sideways}\bl{DVA}\end{sideways}} &~256          &~0.800	&~0.708	&~1.624	&~0.391	&~0.514    \tabularnewline
&~128          &~0.844	&~0.735	&~1.860	&~0.455	&~0.593    \tabularnewline
&~64           &~\bl{0.867}	&~\bl{0.770}	&~\bl{2.100}	&~\bl{0.534}	&~\bl{0.674}    \tabularnewline
&~32           &~\underline{0.859}	&~\underline{0.760}	&~\underline{2.040}	&~\underline{0.527}	&~\underline{0.657}    \tabularnewline
\midrule
\multirow{4}{*}{\begin{sideways}\bl{SalNet}\end{sideways}} &~256          &~0.806	&~0.713	&~1.664	&~0.397	&~0.527    \tabularnewline
&~128          &~0.840	&~0.739	&~1.899	&~0.461	&~0.604    \tabularnewline
&~64           &~\bl{0.868}	&~\bl{0.774}	&~\bl{2.108}	&~\bl{0.533}	&~\bl{0.676}    \tabularnewline
&~32           &~\underline{0.859}	&~\underline{0.756}	&~\underline{2.022}	&~\underline{0.524}	&~\underline{0.651}    \tabularnewline
\midrule
\multirow{4}{*}{\begin{sideways}\bl{SSNet}\end{sideways}}&~256          &~0.807	&~0.712	&~1.650	&~0.403	&~0.523    \tabularnewline\
&~128          &~0.839	&~0.738	&~1.845	&~0.454	&~0.588    \tabularnewline
&~64           &~\bl{0.867}	&~\bl{0.774}	&~\bl{2.100}	&~\bl{0.536}	&~\bl{0.674}    \tabularnewline
&~32           &~\underline{0.858}	&~\underline{0.753}	&~\underline{2.010}	&~\underline{0.528}	&~\underline{0.647}    \tabularnewline
\bottomrule
\end{tabular}
\end{table}

\myPara{Ablation Analysis.} The ablation analyses experiment is conducted to illustrate the contributions of the separate components. To this end, we further implement three ablation models. Two ablation models (Ablation-S and Ablation-T) are  generated without the joint spatiotemporal transfer to distil only the spatial and temporal teacher knowledge, respectively. Similarly, the ablation model Ablation-O is implemented without the separate spatial/temporal distillation so that it can generate the spatiotemporal saliency maps without spatial and temporal teacher knowledge. The performance is presented in \tabref{tab:ablation}, where we can find that the baseline model SKD-DVA-64 achieves the best performance while all the three ablation models have somewhat performance degradation. A possible explanation is that Ablation-S or Ablation-T lacks the sufficient temporal or spatial information in generating aerial video saliency maps, leading to a performance drop. In addition, without the separate spatial/temporal distillation, Ablation-O may have trouble in extracting powerful spatiotemporal cues, resulting in the lowest performance.

\begin{table}[t]
\centering{
\caption{The performance comparisons of the full model SKD-DVA-64 and three baseline models.}
\label{tab:ablation}
\begin{tabular}{l|ccccc}
\toprule
Model&~AUC&~sAUC&~NSS&~SIM&~CC \tabularnewline \midrule
{SKD-DVA-64}&~0.867&~0.770&~2.100&~0.534&~0.674 \tabularnewline
{Ablation-S}&~0.865&~0.769&~2.085&~0.536&~0.669 \tabularnewline
{Ablation-T}&~0.866&~0.762&~2.007&~0.506&~0.647 \tabularnewline
{Ablation-O}&~0.845&~0.751&~1.943&~0.489&~0.621 \tabularnewline
\bottomrule
\end{tabular}
}
\end{table}

\myPara{Efficiency Analysis.} Beyond the effectiveness, the efficiency of our approach is shown in \tabref{tab:performance_AVS1K}. After the step-wisely removing the redundancy in intra-model, data and inter-model, the resulting model contains $0.30$M parameters, leading to a great reduction again $25.07$M, $25.81$M and $41.25$M in DVA, SalNet and STS. Due to this reduction in model parameters as well as resolution, the model memory is greatly reduced, as shown in \tabref{tab:runtime_memory}. Moreover, the inference speeds of our models in different input resolutions are all remarkably improved, as demonstrated in \tabref{tab:runtime_memory}. The inference runtime on the GPU platform can be reduced to $1.414$ms, $0.381$ms, $0.117$ms and $0.035$ms in the resolution of $256\times 256$, $128\times 128$, $64\times 64$ and $32\times 32$, respectively. In particular, SKD-$\mb{T}_s$-32 achieves an extremely fast speed of 28,738 FPS on the GPU platform. These results indicate that our approach provides a practical solution to deploy existing complex deep saliency
models on low-end mobile devices, such as drones.




\begin{table}[t]
\footnotesize
\centering
\caption{Inference time and memory footprint of our approach on high-end GPU (NVIDIA 1080Ti) and low-end CPU (Intel 3.4GHz).}
\label{tab:runtime_memory}
\begin{tabular}{l|c|c|c}
\toprule
\multirow{2}{*}{Model} &~GPU &~CPU&Memory \tabularnewline
                       &~Time~/~\#FPS &~Time~/~\#FPS&footprint (MB) \tabularnewline
\midrule
SKD-$T_s$-256 & 1.414~ms~/~707 & 37.828~ms~/~26.4 &9.24 \tabularnewline
SKD-$T_s$-128 & 0.381~ms~/~2,626 & 9.675~ms~/~103.4 & 2.31 \tabularnewline
SKD-$T_s$-64 & 0.117~ms~/~8,522 & 2.429~ms~/~411.7 & 0.58\tabularnewline
SKD-$T_s$-32 & 0.035~ms~/~28,738 & 0.671~ms~/~1,490.5 & 0.14\tabularnewline
\bottomrule
\end{tabular}
\end{table}

\section{Conclusion} \label{sec:conclusion}
At present, most deep models for dynamic saliency estimation suffer from heavy computational cost and memory footprint, which pose a dilemma for them to be deployed on devices with limited computational capability and memory space. To address this issue, this paper proposes a low-resolution dynamic saliency estimation approach via spatiotemporal knowledge distillation. By step-wisely removing the intra-model, inter-model and data redundancies, a compact and simple saliency model with impressive performance on aerial videos can be established. Experimental results show that the proposed approach outperforms ten state-of-the-art models in estimating visual saliency on aerial videos, while running at an extremely fast speed of 28,738 FPS on the GPU platform. Such a performance means the model can be easily deployed on drones.

In the future work, we will tentatively explore the feasibility of simultaneously distilling the spatial and temporal knowledge from multiple models instead of distilling them separatively. Moreover, the distillation of data will also be explored to further speed up the knowledge distillation from one domain to another one.


%

%

\section*{Acknowledgment}

This work was partially supported by grants from National Natural Science Foundation of China (61672072 \& 61772513), the Beijing Nova Program (Z181100006218063), and Fundamental Research Funds for the Central Universities.

\ifCLASSOPTIONcaptionsoff
  \newpage
\fi



%
%
%
\bibliographystyle{IEEEtran}
\bibliography{egbib}

%

\vspace{-1cm}

\begin{IEEEbiography}[{\includegraphics[width=1in,height=1.25in,clip,keepaspectratio]{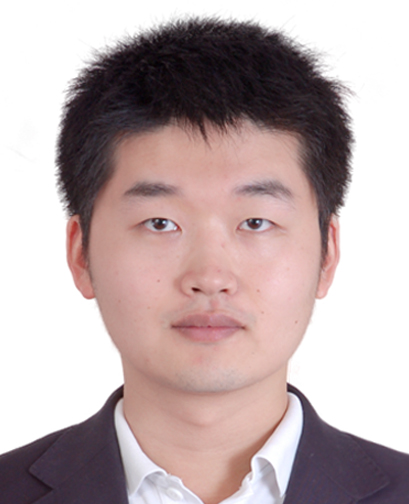}}]{Jia Li} (M'12-SM'15) received the B.E. degree from Tsinghua University in 2005 and the Ph.D. degree from the Institute of Computing Technology, Chinese Academy of Sciences, in 2011. He is currently an associate Professor with the School of Computer Science and Engineering, Beihang University, Beijing, China. Before he joined Beihang University in Jun. 2014, he servered as a researcher in several multimedia groups of Nanyang Technological University, Peking University and Shanda Innovations. He is the author or coauthor of over 50 technical articles in refereed journals and conferences such as TPAMI, IJCV, TIP, CVPR, ICCV and ACM MM. His research interests include computer vision and multimedia big data, especially the learning-based visual content understanding. He is a senior member of IEEE and CCF.
\end{IEEEbiography}

\vspace{-1cm}

\begin{IEEEbiography}[{\includegraphics[width=1in,height=1.25in,clip,keepaspectratio]{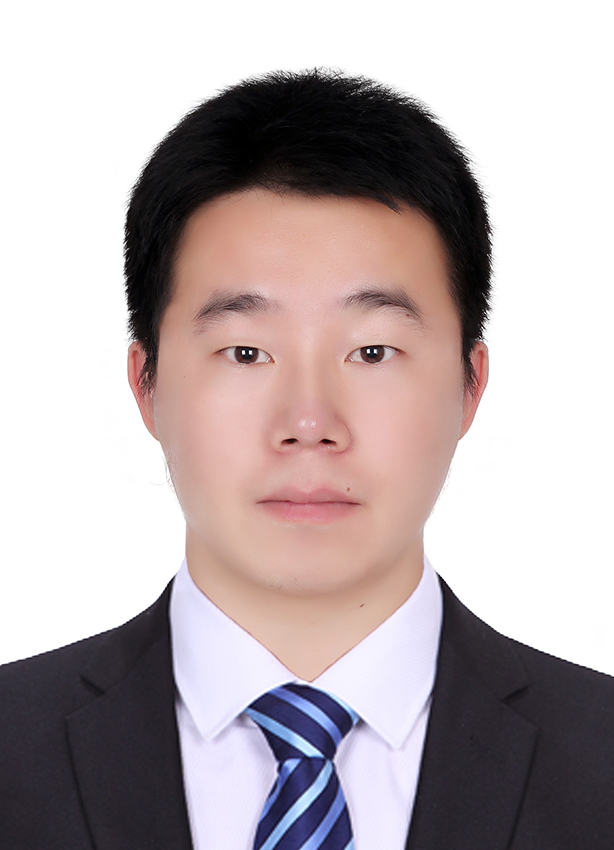}}]{Kui Fu} is currently pursuing the Ph.D. degree with the State Key Laboratory of Virtual Reality Technology and System, School of Computer Science and Engineering, Beihang University. His research interests include computer vision and image understanding.
\end{IEEEbiography}

\vspace{-1cm}

\begin{IEEEbiography}[{\includegraphics[width=1in,height=1.25in,clip,keepaspectratio]{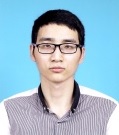}}]{Shengwei Zhao} received his B.S. degree from the School of Mathematics and Statistics in Wuhan University in 2017. He is now a Master student at the Institute of Information Engineering at Chinese Academy of Sciences and the School of Cyber Security at the University of Chinese Academy of Sciences. His major research interests are deep learning and computer vision, especially low-quality image analysis.
\end{IEEEbiography}

\begin{IEEEbiography}[{\includegraphics[width=1in,height=1.25in,clip,keepaspectratio]{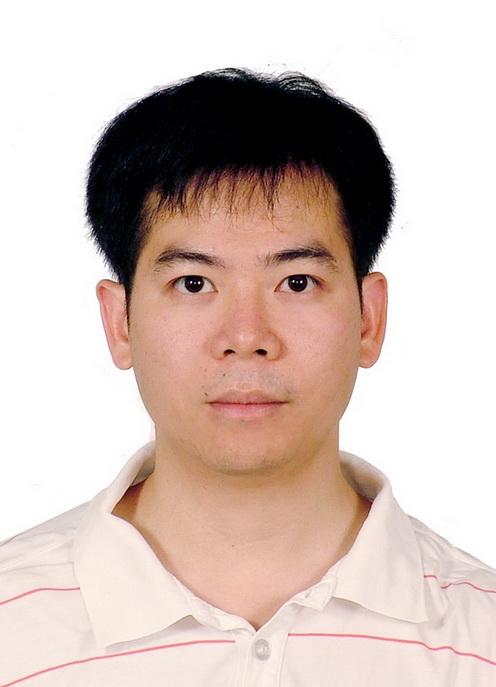}}]{Shiming Ge} (M'13-SM'15) is currently an Associate Professor at Institute of Information Engineering at Chinese Academy of Sciences. Prior to that, he was a senior researcher in ShanDa Innovations, a researcher in Samsung Electronics and Nokia Research Center. He received the B.S. and Ph.D degrees both in Electronic Engineering from the University of Science and Technology of China (USTC) in 2003 and 2008, respectively. His research mainly focuses on computer vision, deep learning and AI security, especially efficient deep learning models and algorithms toward scalable applications.
\end{IEEEbiography}




\end{document}